\documentclass{llncs}
\usepackage{times}

\usepackage{amsmath}
\usepackage{graphicx}
\usepackage[ruled,vlined,linesnumbered]{algorithm2e}
\begin{document}

%\pagestyle{headings}
%In order to omit page numbers and running heads
%please change this line to
\pagestyle{empty}
%and change the first command line too, see above.

\mainmatter

\title{Tackling Air Traffic Conflicts as a Weighted CSP : Experiments with the Lumberjack Method}

\titlerunning{Lumberjack Method}

\author{Nicolas Schmidt, Thomas Chaboud, C\'edric Pralet}%}

\institute{ONERA – The French Aerospace Lab, F-31055, Toulouse, France\\
 \{nicolas.schmidt,thomas.chaboud,cedric.pralet\}@onera.fr}

\maketitle

\begin{abstract}
In this paper, we present an extension to an air traffic conflicts resolution method consisting in generating a large number of trajectories for a set of aircraft, and efficiently selecting the best compatible ones. We propose a multi-man\oe uvre version which encapsulates different conflict-solving algorithms, in particular an original 'smart brute-force' method and the well-known ToulBar2 CSP toolset. Experiments on several benchmarks show that the first one is very efficient on cases involving few aircraft (representative of what actually happens in operations), allowing us to search through a large pool of man\oe uvres and trajectories; however, this method is overtaken by its complexity when the number of aircraft increases to 7 and more. Conversely, within acceptable times, the ToulBar2 toolset can handle conflicts involving more aircraft, but with fewer possible trajectories for each.

\end{abstract}

\section{Introduction}
\subsection{Objective}
\begin{figure}[h]
	\centering
	\includegraphics[width=55mm]{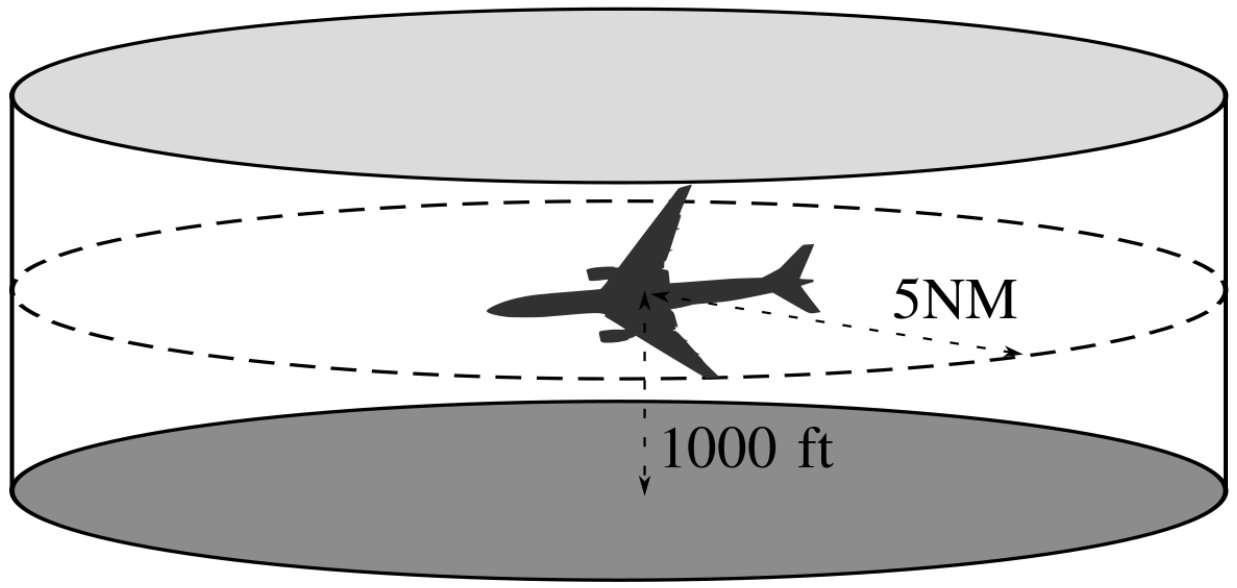}
	\caption{Standard en route separation minima}
	\label{fig:Separation}
\end{figure}

Ensuring aircraft never come closer to each other than regulations demand (cf. Fig.~\ref{fig:Separation}), by ordering changes in speed or temporary deviations from their 
intended course, is the primary task of Air Traffic Control Officers (ATCO). 
A global increasing trend in traffic numbers is forecast, and continental strategic
plans as SESAR (Europe) \cite{SESAR2015} and NextGen (USA) \cite{NASA2012} deem the research and development of automated 
ATC aids necessary to meet this mounting challenge .

Two or more aircraft are involved in a {\it conflict} when their flight paths brings them too close to each other in the near future; upon detection of such a situation, a {\it conflict solver} algorithm computes alternative trajectories ensuring all the aircraft will remain at safe distances before resuming their intended courses. 

A conflict solver's implementation would typically be used in tactical operations, in which time is of the essence: it must suggest its solution to the human ATCO, or -- in a fully automated context -- transmit it directly to the aircraft, in no more than a few tens of seconds. 

Numerous studies have addressed this goal in the last 40 years (see e.g. \cite{KucharYang2000}),
proposing very diverse techniques, among which Linear-/Non Linear Programming or 
various Meta-Heuristics approaches (e.g. \cite{DurandGianazzaGottelandAlliot2015}) are popular. 

The merits of these works are varied as well, considering criteria of computational 
feasibility and of solution trajectories' safety, flyability (respect of the aircraft's flight envelope) and efficiency (e.g. fuel consumption) or their robustness regarding uncertainties. 

However, in our view, not enough stress has been put on the acceptability - mostly from
the ATCO's part - of the proposed conflict solvers. Indeed, an automated aid to critical 
tasks in ATC could not reach operational status without extensive vetting by the 
many actors of Air Traffic Management (ATM), a very conservative world due to the safety 
responsibility pressure. Controllers and pilots would more readily accept in operations computer solutions they can trust (certification) and the algorithms of which they understand and agree upon.  

\subsection{Single man\oe uvre Lumberjack method}

In their 2015 paper \cite {lehouillier2015flexible}, Lehouiller et al. showed that the old method of brute-force searching through a heavily discretised space of simple man\oe uvres was a computationally efficient way to tackle conflict solving. Their algorithm, that we dubbed the {\it Lumberjack}, consists in 3 steps equivalent to the following: 
\begin {itemize}
\item {1. from the input of the aircraft's initial and final positions and speeds, their performances (fuel consumption, maximum and minimum speeds, acceleration, and turn radius), and a given set of man\oe uvres (e.g. 20${^\circ}$ left turn, 10~kt acceleration, etc.), generate for each aircraft the set of every possible trajectory using one man\oe uvre from its starting position before proceeding to the ending one. Each of these paths is affected a cost, e.g. in fuel consumed; }
\item {2. for each pair of aircraft and each pair of their trajectories from step 1, check the paths compatibility: two trajectories are compatible if the aircraft following them keep at least at the minimum separation distance from each other ;}
\item {3. find the minimum cost set of compatible trajectories including one for each aircraft, i.e., solve the Weighted Constraint Satisfaction Problem (WCSP) generated by steps 1 and 2.}
\end {itemize}

\subsection{Pros and cons}
In order to keep the combinatorics manageable, one has to restrict the man\oe uvres set to a well chosen few, which collapses the continuous search space to a small discretised sub-space. The choice of using exactly one man\oe uvre per possible trajectory is a further drastic restriction, though it allows \cite {lehouillier2015flexible} to consider conflict instances involving many (e.g., 15) aircraft. However, we show in this paper that multi-man\oe uvre versions of the algorithm are still efficient enough for cases worse than that occurring in operations, e.g. 5 or 6 aircraft instances. Conflict resolutions comprising several successive man\oe uvres ordered to the aircraft are common in ATC operations, especially when a fast aircraft overtakes a slower one.   

Many of its particularities speak for the Lumberjack method, especially in terms of acceptability: \begin {itemize}
\item {As it is fully deterministic the CPU time is predictably bounded, a definite asset with regard to operational certification.}
\item {In operations, a human ATCO does choose from a few potential new bearings or speed adjustments to order air traffic while solving a conflict. In this sense, the Lumberjack mimics his work much more closely than concurrent techniques; this makes the algorithm itself more easily understandable by ATCOs.} 
\item {If it is used as an automated aid, the conflict solutions it suggests are also easier to validate by a human.}  
\item {Any additional constraint can easily be taken into account, were it e.g. stricter conditions on the trajectories' flyability, avoidance of forbidden areas (Step 1), or separation regulations particular to some aircraft types (Step 2). Similarly, accounting for uncertainties can readily be done as alternative modeling of the trajectories and/or of their separation criteria.}
\item {Other man\oe uvres (e.g. altitude or vertical speed changes) can be added at will in the automated controller's toolbox.}
\item {The method of course outputs an optimal solution within its search space; as we mention below (Section~\ref{sec:FutureWork}), it can also be encapsulated in an iterative algorithm to provide suboptimal any-time solutions.} \\
\end {itemize}

Furthermore, the Lumberjack presents technical and theoretical challenges as different conflict instances generate optimisation problems of varying structures (variable and constraint sets sizes), over which not every WCSP solving technique is efficient. The general method is versatile, and can be applied as a solving pattern for many real-world problems involving collective planning stemming from the aggregation of individual choices; collaborative decision for a group of robots is just one of many examples.

\begin{figure}[!h]
	\centering
	\includegraphics[width=125mm]{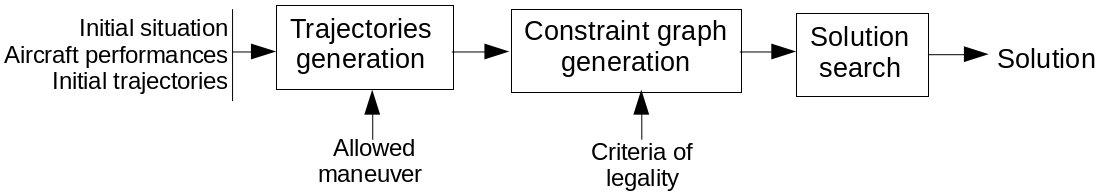}
	\caption{Lumberjack method steps}
	\label{fig1}
\end{figure}

\subsection{Contribution}
In this paper, we present extensions of the single-man\oe uvre Lumberjack to multi-man\oe uvre versions, and encapsulate them in different conflict-solving algorithms. The WCSP of Step 3 above actually is a minimum-cost maximum clique searching problem, for which we devised an original 'smart brute-force' method. It is tested against the well-known ToulBar2 CSP toolset \cite{ToulBar22010} over a number of benchmark sets. This leads us to the question of characterizing conflict problems depending on their graph type and opens the way for machine-learning classification of the instances. 
\subsection{Content}
We focussed more on the method we developed and the resolution of the WCSP, than on the specifics of an operational implementation. The rest of the paper is organised as follows: In Section~\ref{sec:MultiManLJ} we detail the three steps of our multi-man\oe uvre implementation of the Lumberjack method. Section~\ref{sec:SolvingAlgos} presents a 'smart brute force' algorithm (\ref{subs:SBF}), and one using ToulBar2 (\ref{subs:ToulBar2}) to solve the WCSP instance produced. Section~\ref{sec:Expe} describes our experiments to compare these methods and their results, while Section~\ref{sec:FutureWork} is devoted to the follow-up research we envision.

\section{Multi-man\oe uvre Lumberjack method}
\label{sec:MultiManLJ}
We expand here upon the three steps of the general Lumberjack algorithm, adapted to the multi-man\oe uvre trajectory generation.

\subsection*{Step 1 - trajectory generation}

This step depends on a discretisation of the man\oe uvres space. Similarly to the practice of a human ATCO, the man\oe uvres selected are simple changes of bearing or speed with a few discrete values (see below). The discretisation's granularity, which tells how many different choices one has in each order type (turn, speed) is an input parameter.

%- discretisation temporelle\\
The discretisation is done regarding time as well: The resolution period is broken down into equal time segments at the beginning of which one man\oe uvre can start. The number of these segments is another input parameter. 

%- manoeuvres consid\'er\'es\\
In the implementation of the method used for this paper, we chose the set of orders to include:
%L'ensemble des manoeuvres consid\'er\'e est le suivant :
\begin{itemize}
\item Turn: change of bearing, to the left or right, of a given number of degrees using a step of 20, 10, 5 or 2$^{\circ}$ depending on the granularity.
\item Speed : acceleration or deceleration of a percentage of the current aircraft's speed, using a step of 20, 10, 5 or 2\%. Actually, a human ATCO would rather request round number objective speeds in such cases, but this choice was slightly more convenient in our implementation. 
\item Turn-and-Speed: Both the above man\oe uvres executed on the same time segment. As ATCOs usually avoid giving such complex orders to a pilot, we only use that type of control when the period is split into very few segments. 
\item Do-Nothing: The aircraft keeps its current speed and bearing for the duration of the time segment.
\item Straight-to-end: Clearance to go directly to the final position, adjusting both bearing and speed as needed.
\end{itemize}
The first three are called {\it man\oe uvre orders}, while the last two are {\it non-man\oe uvre orders}. 
Every trajectory generated ends with a Straight-to-end type order, which can happen before the last time segment.

%- l\'egalit\'e\\
The trajectories generation routine checks their {\it legality}, i.e. whether they respect the aircraft's performance, avoid potential forbidden zones, and reach the final position in due time.\\%- calcul de couts\\
A cost function allocates a value to each individual trajectory; we used a simple fuel consumption model, with parameters depending on the aircraft's performance category.

%- param\`etres : discr tempo, limite nb man, granularit\'e\\
The choice of the discretisation parameters allows one to modify: 
\begin{itemize}
\item the number of time segments in the resolution period, hence the depth of the tree representing the set of trajectories;
\item the number of man\oe uvre orders (other than some 'Do-nothing' ones and the final clearance), so as to limit the length of the complete order sequence (description of the trajectory planned) given to each pilot by the ATC;
\item the granularity of the man\oe uvre space discretisation, hence the number of possible man\oe uvres at the start of each segment.
\end{itemize}

For the sake of clarity, let us note that the same conflict problem instance may be transformed into several optimisation problems which can be quite different in size and/or structure, depending on the 3 parameters above.

\begin{figure}[!h]
	\centering
	\includegraphics[width=40mm]{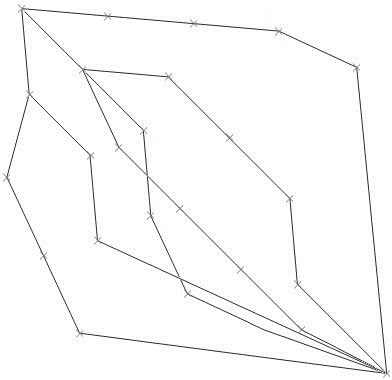}
	\caption{Potential trajectories going from upper left to lower right.}
	\label{fig3}
\end{figure}

Fig.~\ref{fig3} illustrates the result of this phase. The input parameters used are: 6 segments in the resolution period; available man\oe uvres: Turn -40$^{\circ}$, -20$^{\circ}$, 20$^{\circ}$, or 40$^{\circ}$, Do-nothing, Straight-to-end; at most 3 man\oe uvre orders allowed. Just 6 randomly chosen trajectories among the 600 generated are represented; they join the starting position in the upper left to the lower right. Detailing the uppermost path, it includes a 40$^{\circ}$ Turn on segment~1, a -20$^{\circ}$ Turn on segment~4, and a Straight-to-end on segment~5.\\

Given $p$ the number of time segments, $m$ the maximum number of man\oe uvre orders in the sequence, and $n$ the number of possible man\oe uvres, the algorithm generates at most    

\begin{center}
${\dbinom{m}{p}}~n^m $	
\end{center} 
trajectories, including illegal ones. Given the settings we chose, our experiments usually ranged between about $500$ and $25000$ trajectories per aircraft.

In operations, the computing time is critical; in order to keep such large numbers of objects still tractable, we chose simple models of the man\oe uvres: a change of bearing is represented as a point-turn between two straight segments, and a change of speed is made at a constant acceleration or deceleration until reaching the requested one. Higher fidelity representations of the trajectories is not deemed necessary in this study, in part because the resolution synthesises an order given by an ATCO to pilots, the reaction delay of whose is not known for certain. The approximation errors, both in real day-to-day operations and in our experiments, are generally compensated by additional safety margins over the separation minima e.g. of Fig.~\ref{fig:Separation}.

\subsection*{Step 2 - constraints graph generation}
The goal of this step is to determine the compatibility of each pair of trajectory for each pair of aircraft, defined as the respect of the separation distance between each of their segments. This compatibility is a hard constraint in the WCSP generated.\\ 

This step's complexity is quadratic over the maximum number of trajectories per aircraft: with $t_x$ and $t_y$ respectively the number of trajectories generated for aircraft $x$ and $y$, checking their one-to-one compatibility asks for about $t_x \times t_y$ costly operations. Even considering that the distance computations, themselves at worst quadratic over the number of segments per trajectory, can be reduced to $1~\mu s$, executing Step 2 for e.g. 3 aircraft having 3000 potential trajectories each takes $3*3000^2*1~\mu s=27~s$.\\

Our 'smart brute force' way of solving the WCSP (cf. \ref{subs:SBF}) owes much of its comparative efficiency by its gains over this phase of the basic method as it only checks the trajectories' compatibility on-the-fly, while looking for the minimum-cost maximum clique.

%-fonction de calculs du respect de la distance de s\'ecurit\'e? ou juste le principe : segment par segment?\\

\subsection*{Step 3 - Searching for solutions}

The problem can either be solved as the WCSP formalised below, or -- as \cite{lehouillier2015flexible} showed -- by finding a minimum cost maximum clique in an $n$-partite graph, with $n$ the number of aircraft: The graph nodes are the trajectories valued by their cost, two nodes being adjacent iff the corresponding trajectories are compatible. Any $n$-clique in this graph is by definition a set of aircraft trajectories respecting separation minima, and the minimum cost $n$-cliques are optimal resolutions of the conflict represented, within the search space.\\ 

These two approaches allow us to devise two conflict-solving algorithms, one using ToulBar2 on the WCSP (cf. \ref{subs:ToulBar2}) and the other a 'smart brute-force' search for the best $n$-clique in this graph (\ref{subs:SBF}), which our experiments prove more efficient on some categories of instances. 
Furthermore, we added a fast greedy non-optimal method (\ref{subs:GreedyMethod}) in order either to quickly provide ToulBar2 with an a priori upper bound (which improves markedly its performances), or to solve instances too large to be tractable by the optimal methods. 

\section{Solving algorithms}\label{sec:SolvingAlgos}
\subsection{Formalisation of the WCSP}
Once every possible legal trajectory for each aircraft has been generated, its cost computed, and the compatibilities set, the resolution is that of a WCSP in which the variables are the aircraft, the variables' domains are the sets of trajectories, the trajectories' cost are soft constraints, the compatibilities are hard constraints, and the solution minimises costs.\\

Formally, the WCSP instance is given by a tuple $P=(\top, \mathcal{X}, \mathcal{D}, \mathcal{C})$, with 
\begin{itemize}
	\item $\mathcal{X}=\{x_1,...,x_n\}$ a set of variables;
	\item $D_i \in \mathcal{D}$ the associated finite domain of $x_i$;
	\item $\mathcal{C}=\{c_1,...,c_m\}$ a set of constraints;
	\item $\top$ the upper bound of the problem.
\end{itemize}
A tuple $t$ is an assignment to a set of variables such that $Var(t) \subseteq \mathcal{X}$, and the projection of a tuple $t$ over a set of variable $X$ is denoted $t \downarrow_X$. Each constraint $c$ associates any instantiation of $Var(c)$ to a cost $\phi=[0,\top]$. For readability reasons, we will note $c(t)$, instead of $c(t\downarrow_{Var(c)})$, the cost associated to a tuple $t$ by the constraint $c$.\\
A constraint $c$ is called \textit{soft} if $\exists t$ such that $0 < c (t) < \top$; conversely, a constraint $c$ is called \textit{hard} if $\forall t$, we have $c (t)=\{0, \top\}$.\\
The cost of a tuple, denoted $\mathcal{V}(t)$, is the sum over all applicable costs $c(t)$ such that \mbox{$\forall c$, $Var(c)\subseteq Var(t)$}. A tuple $t$ is a solution of the problem iff $Var(t)=\mathcal{X}$ and $\mathcal{V}(t)<\top$.\\

The aircraft are the variables in $\mathcal{X}$, and each one's $d_i$ trajectories are the elements of $D_i=\{1,...,d_i\}$. The set of constraints is made of unary soft and binary hard ones, respectively corresponding to the cost of each trajectory, and to the compatibility status of each pair of trajectories.\\
The $\top$ is the cost of the best known solution plus one, and if no solution is known, it is the sum of the costs of the worst trajectory of every aircraft.

\subsection{Solving with ToulBar2}\label{subs:ToulBar2}
This WCS problem is simply written in a .wcsp formatted file for ToulBar2 to solve. However, just generating the instance is quite costly, in terms of both CPU time and file size because the constraints must be exhaustively enumerated while the variables' domains can be very large (e.g. 25000 trajectories per aircraft). The method presented below saves up on that computing cost.

\subsection{'Smart Brute Force' (SBF) maximum clique searching}\label{subs:SBF}
In the problem graph defined as in Step~3 of Section~\ref{sec:MultiManLJ} above, this algorithm searches for solutions as $n$-cliques. Its interest lies in that it only computes the part of the constraint graph (Step~2) which is strictly necessary to find the minimum cost of such cliques: The trajectories of each aircraft are sorted by order of increasing cost, and the search is organised such that the first clique found is that of minimal cost, in the following way (cf. also the formalised version, Algorithm~\ref{mcMC}).

By construction, the graph is made of the $n$ independent sub-graphs representing the sets of each aircraft's trajectory. Tuples of trajectories are examined in turn, starting with the single made of the first (less costly) trajectory of the first aircraft. On each step, we select the $i$-uple with minimal cost (sum of the costs of its elements) in a list. We add in the list the $i$-uple corresponding to the following node of the (ordered) sub-graph~$i$ and, if it is a clique (its elements are compatible), we also add to the list the $i$+$1$-uple obtained by adding the first trajectory from the ($i+1$)\textsuperscript{th} sub-graph. As a side note, using a binary min-heap is an efficient way of storing and managing the $n$-uples. \\

\begin{algorithm}[!h]\label{mcMC}
	\caption{$\mathsf{MinCostMaxClique}$}
	\SetKwSty{textrm}
	\SetKwInOut{Struct}{With }
	\SetKwInOut{Output}{output }
	\DontPrintSemicolon
	
	\Struct{- $p$, the number of aircraft\\
		- Vect,\,a\,$struct$\,representing\,a\,tuple\,of\,trajectories,\,and\,composed\,of\,an\,array\,$tab$\,of\\size $p$, containing the id of nodes from the $p$-partite graph; an $index$; and a $cost$\\
		- H, a binary min heap of Vect sorted by their $cost$}
	\Output{either a Vect $v$ containing the optimal solution, or "\textit{no solution}"}
	\BlankLine
	Vect $v, v'$;\\
	$v.tab[0~..~p$-$1]:=-1$;~ $v.index:=0$;\\
	$v.tab[v.index]$++;\\
	H$.insert(v)$;
	
	\While {H isn't empty}
	{
		$v:=$H$.extract\_top$;\\
		\label{legal}\uIf {$v$ is legal}{
			\uIf {$v$ is solution}{return $v$;}
			$v':=v$;\\
			$v'.index$++; $v'~ .tab[index]$++;\\
			H.$insert(v')$; 
			%H.$insert(v'.tab[index]$++);
		}
		$v.tab[index]$++;\\
		H.$insert(v)$;
	}
	
	\Return "\textit{no solution}";
\end{algorithm}

The graph edges are not set a priori, but on-the-fly of this procedure, on Line~\ref{legal} of Algorithm~\ref{mcMC}, only on the relevant trajectories; this way, only the necessary part of the adjacency matrix is constructed incrementally. 

Thus, this method is especially powerful on 'simple' problems, i.e. those that involve a small number of aircraft, and in which the solutions do not deviate much from a direct path between the end positions; remarkably, such cases are by far the most frequently occurring in operations. In such instances, its efficiency allows to deal with large numbers of trajectories per aircraft, and hence to increase the resolution fidelity via the granularity parameter.\\

%- limites
 However, the SBF algorithm becomes impractical as the exponential complexity grows, e.g. on 7 or 8 aircraft instances, that favour the structural tricks of higher-level algorithms like ToulBar2. As mentioned, though, these cases never appear in operations because they are prevented by pre-tactical or ATFCM (Air Traffic Flow and Capacity Management) regulations, and will still be even if the traffic increases. 

\subsection{Non-optimal greedy method}\label{subs:GreedyMethod}
In addition to the two resolution methods presented above, we present a greedy sequential algorithm that can solve the problem quickly while forgoing the optimality: WCSP solvers as ToulBar2 greatly benefit of the input of an a priori upper bound over the solution cost; it is also a way to reduce the average computing time of Step~2 of Section~\ref{sec:MultiManLJ}, that of the trajectories compatibility, which becomes unnecessary if the cost involved is already too large. Furthermore, a fast non-optimal greedy method is the last resort if the instance considered is too complex for better algorithms.\\

The $n$ aircraft are considered sequentially. In each step \mbox{$i\in\{1,...,n\}$}, we add to the solution $i$-uple the minimal cost trajectory of aircraft $i$, that is compatible with the first $i-1$ ones. This way, complexity of Steps 2 and 3 of Section~\ref{sec:MultiManLJ} becomes linear in the total number of trajectories; in many-aircraft instances, perhaps from a longer-term operational context (i.e., pre-tactical or strategic), one can thus find a solution without reducing the granularity too much. Conversely, one can use the greedy algorithm on a higher granularity setting than with the optimal methods and sometimes find better solutions to the same conflict problem (cf. Section~\ref{subs:Beyond}).

\section{Experimentation}\label{sec:Expe}
\subsection{benchmarks and protocol}\label{subs:Benchmarks}
%-presentation des bench : cas d'ecole : rond point, croisement, autres?\\

\begin{figure}[h]
	\centering
	\includegraphics[width=75mm]{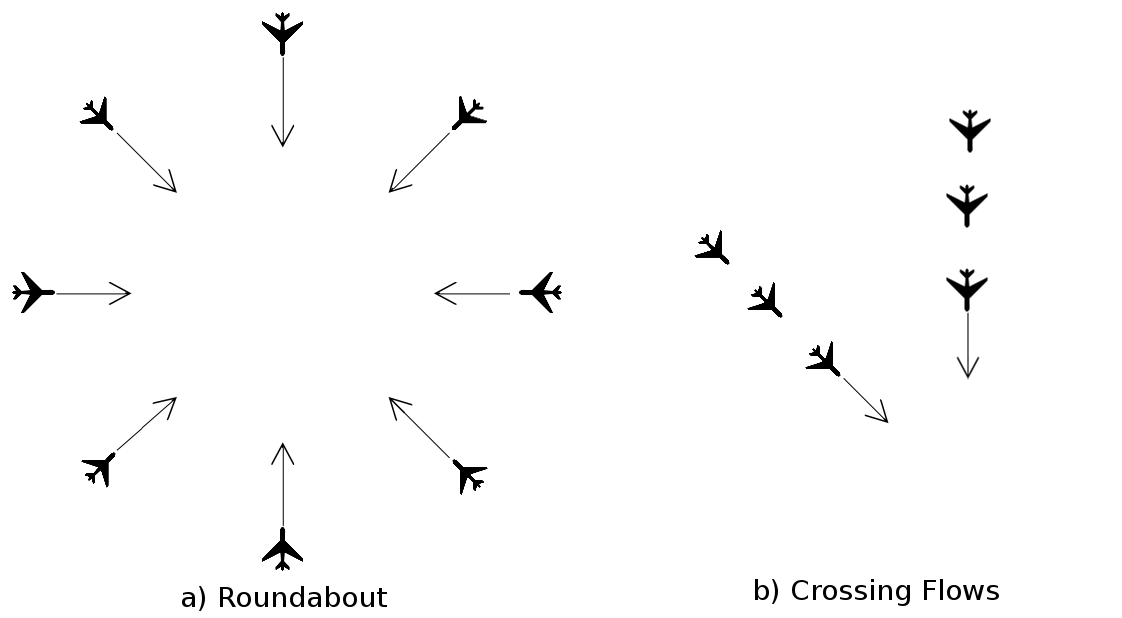}
	\caption{Roundabout: The aircraft come from all directions, their initial trajectories passing through a common central point; Crossing Flows: The aircraft are grouped on two lines that intersect.}
	\label{fig:Bench}
\end{figure}

We evaluate and compare the different methods presented above over two benchmark types used before in the conflict-solving literature (Fig.\ref{fig:Bench}). Their instances share the same geometry, with randomised initial and final positions and speeds, but differ in number of aircraft involved; the aircraft's performances are randomised as well.\\
 
%Nous proposons, afin de comparer les differentes approches, plusieurs benchmarcks reprenant des cas d'ecole.\\

The Crossing Flows case is simpler, mostly because every aircraft does not interfere with each other as in the Roundabout case; also, it usually requires only minor path adjustments. It is more representative of an operational problem as well: such situations happen at the crossing of major airways. Conversely, the Roundabout is theoretically impossible in reality, because aircraft flying in opposite directions do so on separate flight levels, according to airspace rules. Nonetheless, its high complexity in terms of interactions between all aircraft and the large changes of trajectory its solutions include make it a challenging benchmark.\\

So as to make our tests on these benchmarks more statistically representative, we execute each of them in a loop, e.g. 100 times, that differ by some random noise added in the data (initial and final positions and speeds); the result considered is the average performance over the loop. We analyse each method's computing time and, with a 60 seconds time-out, its success rate. Note that this time-out is chosen for the sake of more detailed comparisons, the actual goal times being closer to 20~s.

We also studied conflict situations from real day-to-day operations, but they nearly all involve 2 aircraft only, very rarely 3 and exceptionally 4. The results obtained on them are similar to those we got on small instances of the Crossing benchmark . 

\subsection{Results}\label{subs:Results}
%-cadre experimental\\
All tests were performed on a computer equipped with the following hardware: Intel Core~2 Duo processor, 3.00~GHz, 8~GB RAM. All algorithms were implemented in C++ and compiled by GCC~4.4.7. We used the Toulbar2.0.9.6.0 version of Toulbar, set on default options which so far give us the bests results.\\

\begin{figure}[h]
\centering
\includegraphics[width=75mm]{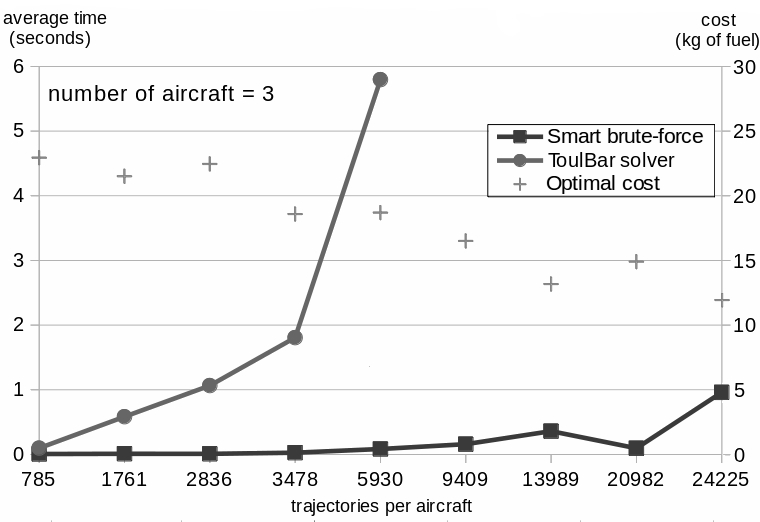}
\caption{Average solving time over 100 tests of the SBF (squares) and Toulbar2 (circles) approaches for instances of a Roundabout with 3 aircraft, function of the number of generated trajectories per aircraft. Crosses represent the average cost of the optimal solution.}
\label{fig:graph1}
\end{figure}

Figure~\ref{fig:graph1} illustrates the WCSP/ToulBar2 and SBF methods' behaviour on a 3-aircraft Roundabout benchmark. It displays the increase of computing time along with that of the number of trajectories per aircraft. One can see that SBF always gives the optimal solution in an acceptable time, while ToulBar2 meets trouble when the constraint set grows too much; its size is related to the square of the number of trajectories per aircraft. When this parameter becomes larger than 9000 the failures are actually due to memory overflows, though the curve's aspect shows time-outs would have been likely. Being able to deal with a large number of trajectories per aircraft is important, as it generally affects the solutions' quality, shown as crosses in the figure. SBF thus has decisive merits over such small conflict instances.\\

It may come as a surprise that the optimal cost does not decrease monotonically (e.g., here it is higher with 20982 trajectories per aircraft than with 13989), but this is due to a discretisation effect, which can also be seen to impact SBF's average computing time:   

\begin{center} \begin{table*}[h]\caption{Setting of the discretisation parameters in Fig.~\ref{fig:graph1}}\label{table1}
		\begin{center}
		\begin{tabular}[b]{|c|c|c|c|c|}
			\hline
			\# segments  & \# man. orders & granularity & \# man\oe uvres & \# trajectories \tabularnewline
			\hline
			4 & 2 & 2  & 17 & 785.47  
			\tabularnewline
			5 & 2 & 2 & 17 & 1761.61  
			\tabularnewline
			4 & 3 & 2 & 17  & 2836.8 
			\tabularnewline
			6 & 2 & 2 & 17 & 3478.37  
			\tabularnewline
			7 & 2 & 2  & 17 & 5930.66 
			\tabularnewline
			8 & 2 & 2 & 17 & 9409.97  
			\tabularnewline
			6 & 2 & 3 & 33 & 13989.03  
			\tabularnewline
			4 & 3 & 3 & 33  & 20982.39        
			\tabularnewline
			7 & 2 & 3  & 33 & 24225.66 
			\tabularnewline
			\hline
		\end{tabular}\end{center}
	\end{table*}
\end{center}

Producing more trajectories per aircraft, on a given conflict problem instance, is done by tweaking the 3 discretisation parameters: number of time segments, number of man\oe uvre orders allowed, and the granularity which directly translates into the number of available man\oe uvres. However, these settings are only chosen among small integral values; this produces optimisation problems differing both in size of their constraint set and in their number of variables. Neither of these descriptions, nor any simple combination of them, gives the problems difficulty a total order; we chose the latter (variables set size) to present the tests in Fig.~\ref{fig:graph1}, as the least bad choice.

As a side note, "number of trajectories" decimal values may seem strange as well, as fixing the 3 discretisation parameters of course generates same-sized sets of potential trajectories, even taking into account the noise added to the data in each run; but the number of {\it legal} trajectories varies, and it is what is averaged in Table~\ref{table1}'s last column.\\

\begin{figure}[h]
	\centering
	\includegraphics[width=75mm]{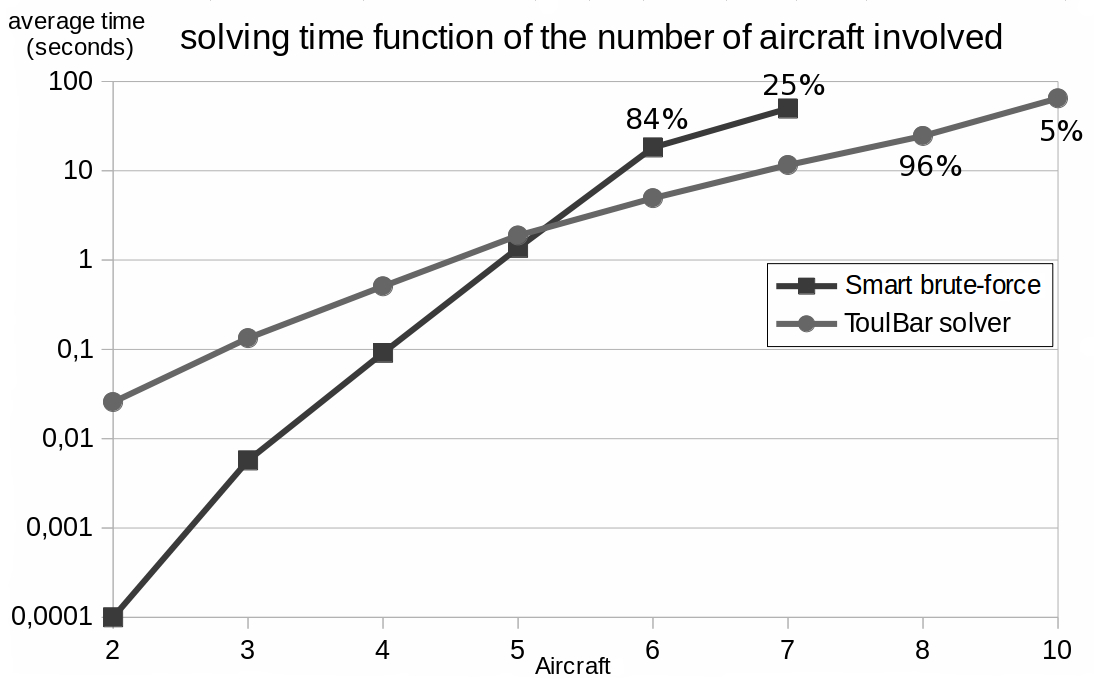}
	\caption{Average solving time over 100 tests with statistical noise, of SBF (squares) and Toulbar2 (circles) approaches for instances of a Roundabout with approx. 850 trajectories per aircraft, function of the number of aircraft involved. When a method does not always succeed before the 60~s time-out, its success rate is indicated.}
	\label{fig:graph2}
\end{figure}

Figure~\ref{fig:graph2} shows how the computing times behave on instances of a Roundabout conflict involving an increasing number of aircraft, while the number of trajectories per aircraft is constant.
The left part of the figure confirms our  above analysis: SBF is more efficient than ToulBar2 on small conflict problem instances, i.e. involving 5 or fewer aircraft. The new information on the right part is that ToulBar2 clearly overtakes SBF's performance from 6-aircraft instances on, nearly always solving the problem (within 60~s) up to 8-aircraft instances, where its opponent fails.  

\begin{center} \begin{table*}[!h]\caption{Evolution of solving times on the Crossing Flow benchmark for 2, 4, 6 and 8 aircraft, and different input parameters}\label{table2}
		\begin{tabular}[b]{|c|c|c|c|c|c|c|c|}
			\hline
			Number of  & parameters & trajectories & \multicolumn{3}{c|}{ToulBar method} & SBF & optimal solution \tabularnewline
			aircraft & ($p$;$n$;$m$) & per aircraft & generation (s) & solving (s) & total (s) &  (s) & (kg of fuel) \tabularnewline
			\hline
			2 & (2;2;33) & 929  & 0.01 & 0.03 & \textbf{0.04} & \textbf{0.01} & 19.68 \tabularnewline
			2 & (4;2;17) & 1924 & 0.06 & 0.14 & \textbf{0.2 } & \textbf{0.01} & 17.46 \tabularnewline
			2 & (5;2;17) & 3694 & 0.2  & 0.54 & \textbf{0.74} & \textbf{0.02} & 18.32 \tabularnewline
			2 & (7;2;17) & 9809 & 1.36 & m-o  & \textbf{-   } & \textbf{0.17} & 17.28 \tabularnewline
			\hline
			4 & (2;2;33) & 929  & 0.31 & 0.24 & \textbf{0.55} & \textbf{0.14} & 42.15 \tabularnewline
			4 & (4;2;17) & 1924 & 1.77 & 1.1  & \textbf{2.87} & \textbf{0.33} & 36.85 \tabularnewline
			4 & (5;2;17) & 3694 & 9.43 & 5.48 & \textbf{14.91}& \textbf{10.26}& 39.75 \tabularnewline
			4 & (7;2;17) & 9809 & t-o  & -    & \textbf{-   } & \textbf{t-o}  & -     \tabularnewline
			\hline
			6 & (2;2;33) & 929  & 2.01 & 0.72 & \textbf{2.73} & \textbf{6.6}  & 67.57 \tabularnewline
			6 & (4;2;17) & 1924 & 9.97 & 3.29 & \textbf{13.26}& \textbf{49.44}& 62.30 \tabularnewline
			6 & (5;2;17) & 3694 & t-o  & -    & \textbf{-   } & \textbf{t-o}  & -     \tabularnewline
			6 & (7;2;17) & 9809 & t-o  & -    & \textbf{-   } & \textbf{t-o}  & -     \tabularnewline
			\hline
			8 & (2;2;33) & 929  & 6.41 & 1.82 & \textbf{8.23} & \textbf{t-o}  & 93.22 \tabularnewline
			8 & (4;2;17) & 1924 & 29.07& 7.62 & \textbf{36.69}& \textbf{t-o}  & 87.69 \tabularnewline
			8 & (5;2;17) & 3694 & t-o  & -    & \textbf{-   } & \textbf{t-o}  & -     \tabularnewline
			8 & (7;2;17) & 9809 & t-o  & -    & \textbf{-   } & \textbf{t-o}  & -     \tabularnewline
			\hline
		\end{tabular}
	\end{table*}
\end{center}

Table~\ref{table2} displays the results of both methods over the Crossing Flows benchmark. Regarding the ToulBar2 part, we distinguished the constraint set generation and writing time, from the solving time.
($p$;$n$;$m$) stands for the discretisation parameters: $p$ is the number of time segments, $m$ the  number of man\oe uvres allowed, and $n$ the number of available man\oe uvres; "t-o" stands for time-out (60~s), "m-o" for a memory overflow, "-" for no result.

The same remarks made regarding Fig.~\ref{fig:graph1} and Table~\ref{table1} still hold: in a few cases, the solution becomes worse whereas e.g. the number of segments increases.

As a further confirmation of our previous results, SBF solves 5 or fewer aircraft instances faster than ToulBar2, while the latter is better or the only one succeeding on 6 and 8 aircraft conflicts. Again, more than 9000-trajectory instances prove intractable for ToulBar2; one could argue that this is due to hardware limitations, but the issue is intrinsic (the constraints set is large) and would reappear at higher thresholds on bigger computers -- eventually, complexity always wins.\\

More generally, the different behaviours of these two methods are clear in our examples: representing one of our problems as the trajectory graph with compatibility edges, as described in Step~3 of Section~\ref{sec:MultiManLJ}, ToulBar2 is efficient on $n$-partite graphs as long as the $n$ independent sub-graphs themselves remain small enough. On the other hand, SBF handles large sub-graphs better but only if they are not too numerous, i.e. if $n$ is small.

\subsection{Beyond the limits}\label{subs:Beyond}
When the number of aircraft involved in a conflict is too high, it can be interesting to use a non-optimal greedy method such as the one presented in Section~\ref{subs:GreedyMethod} above. Indeed, in these cases, optimal solution search methods like ToulBar2 or SBF could only hope to succeed in an acceptable time on very low discretisation parameters, whereas the greedy algorithm can work on higher parameters and may find a non-optimal solution that beats the low parameters optimal ones.\\

We put that idea in practice on a 20 aircraft Roundabout conflict instance. With discretisation parameters generating 82 trajectories per aircraft, ToulBar2 took 12~s to find a solution; on settings producing 290 trajectories per aircraft, it found a 15\% better solution in 30~s. 

On the other side, setting the parameters to generate about 10000 trajectories per aircraft, the greedy method output a (non-optimal) solution 40\% better than that of ToulBar2 in just 6~s.\\

We observed that the greedy algorithm becomes more efficient when around 16 aircraft are involved in a conflict. Of course this case is far beyond the worst situations occurring in operations, but the general technique is likely to be of use in some of the many possible applications of the methods described in this paper. 

\section{Ongoing and future work}\label{sec:FutureWork}
 With the insight of the results presented above, we have developed an iterative process using the SBF solver in a loop: the discretisation parameters are increased step after step, enlarging the trajectory search space; each iteration provides an upper bound benefiting the next. This technique has the added advantage to provide any-time solutions, an important feature in time-critical operations. We are still studying the parameters exploration strategy, with the difficulties highlighted by the above remark about Fig.~\ref{fig:graph1} and Table~\ref{table1}; parallelising it could be a viable option.
 
The differing behaviours and efficiencies of the WCSP/ToulBar2 method on the one hand, and SBF on the other, lead us to attempt a precise classification of the conflict instances into categories on which each algorithm would work best. 

Besides, for now, our 'granularity' parameter refines all the man\oe uvre order types (e.g., Turn, Speed, Climb...) equally. One suspects that more specialised tool boxes of man\oe uvres to build the potential trajectories would work best depending on some typology of the conflict instances (e.g. Crossing at a given angle range, Take-over, Level crossing...). This is another classification we intend to work on, with the help of e.g. Principal Component Analysis and/or diverse machine learning techniques.\\

We noted the versatility of methods akin to the Lumberjack, lending them to applications in many other fields; of course, our previous work on ATC would more easily adapt to germane ATM tasks, like pre-tactical/strategic traffic flow management. Even remaining in the ATC world, the challenges of Terminal and Approach areas differ enough from the en route type control we concentrated on in this paper to deserve their own studies.

\section{Conclusion}\label{Conclusion}
We showed that our extension of the single-man\oe uvre Lumberjack to multi-man\oe uvre versions, along with the algorithms we devised to solve the optimisation problems, seems a promising candidate towards implementation in operational ATC tools, keeping in mind that our analysis is only a first step in the protracted ATM validation process. As the different optimisation methods' behaviours complete each other to extend the spectrum of tractable conflict instances, such tools would preferably include several algorithms and quickly select that which is best suited to the problem at hand.

Our research prospects are focussed both on ways to classify conflicts so as to choose the best method and tweak its parameters, e.g. via machine learning techniques, and on diversifying our algorithms' applications towards e.g. strategic/pre-tactical traffic management or Terminal Area ATC.

%\section*{Acknowledgments}

%acknowledgments

\bibliographystyle{plain} %splncs}
\bibliography{bibCP}

\end{document}